\documentclass{article}
\usepackage[utf8]{inputenc}
\usepackage{iclr2027_conference,times}

\usepackage{amsmath,amsfonts,bm}

\def\eqref#1{equation~\ref{#1}}

\def\1{\bm{1}}

\DeclareMathAlphabet{\mathsfit}{\encodingdefault}{\sfdefault}{m}{sl}
\SetMathAlphabet{\mathsfit}{bold}{\encodingdefault}{\sfdefault}{bx}{n}

\usepackage{wrapfig}
\usepackage{amsmath}
\usepackage{amssymb}
\usepackage{booktabs}
\usepackage{graphicx}
\usepackage{algorithm}
\usepackage{algpseudocode}
\usepackage{hyperref}
\usepackage{url}

\title{Bootstrapped On-Policy Distillation}

\author{
Yue Wu$^{1}$\thanks{Equal contribution.},
Xueyu Zhou$^{1}$\footnotemark[1],
Yangrong Hu$^{1}$,
Jian Huang$^{1,2}$\thanks{Corresponding author: Jian Huang.} \\
\textnormal{$^1$Department of Data Science and Artificial Intelligence} \\
\textnormal{$^2$Department of Applied Mathematics} \\
\textnormal{The Hong Kong Polytechnic University} \\
\texttt{\{yue0301.wu,xueyu.zhou,yangrong.hu\}@connect.polyu.hk} \\
\texttt{j.huang@polyu.edu.hk}
}

\iclrfinalcopy

\begin{document}

\maketitle

\begingroup
\renewcommand{\thefootnote}{}
\footnotetext{Code is available at
\url{https://github.com/YueWu0301/BOPD}.}
\endgroup

\lhead{Preprint}

\begin{abstract}
On-policy distillation (OPD) improves language models by distilling teacher predictions on student-generated trajectories, while on-policy self-distillation (OPSD) removes the need for an external teacher by using a reference-conditioned copy of the model as the teacher.
Despite its great early improvements, we find that OPSD does not remain consistently effective under continued training and can eventually suffer substantial performance degradation.
We study this training behavior and link the degradation to a growing mismatch between the evolving reference-free student and the frozen privileged teacher.
In the early stage of OPSD, the privileged teacher remains closely aligned with the student and provides useful guidance for improving the student's generation trajectories. However, as training progresses, the growing distributional mismatch between the privileged teacher and reference-free student can cause the teacher to provide misleading supervision, ultimately degrading student performance.
This phenomenon motivates us to first leverage privileged reference supervision to acquire useful knowledge and then transfer it to a fresh student.
We introduce Bootstrapped On-Policy Distillation (BOPD), a two-stage framework that first uses OPSD to internalize task-relevant information from reference-conditioned supervision into an intermediate model, and then distills this reference-free intermediate model into a fresh copy of the initial model.
Experiments across models show that BOPD improves mathematical reasoning and code generation beyond both the initial models and the constructed OPSD teachers.
These results suggest that privileged reference supervision can be more effectively exploited by first transferring its task-relevant signal into a task-adapted model and then distilling from that model without reference conditioning.

\end{abstract}

\section{Introduction}
\label{sec:introduction}
\begin{wrapfigure}{r}{0.4\textwidth}
    \centering
    \vspace{-10pt}
    \includegraphics[width=0.4\textwidth]{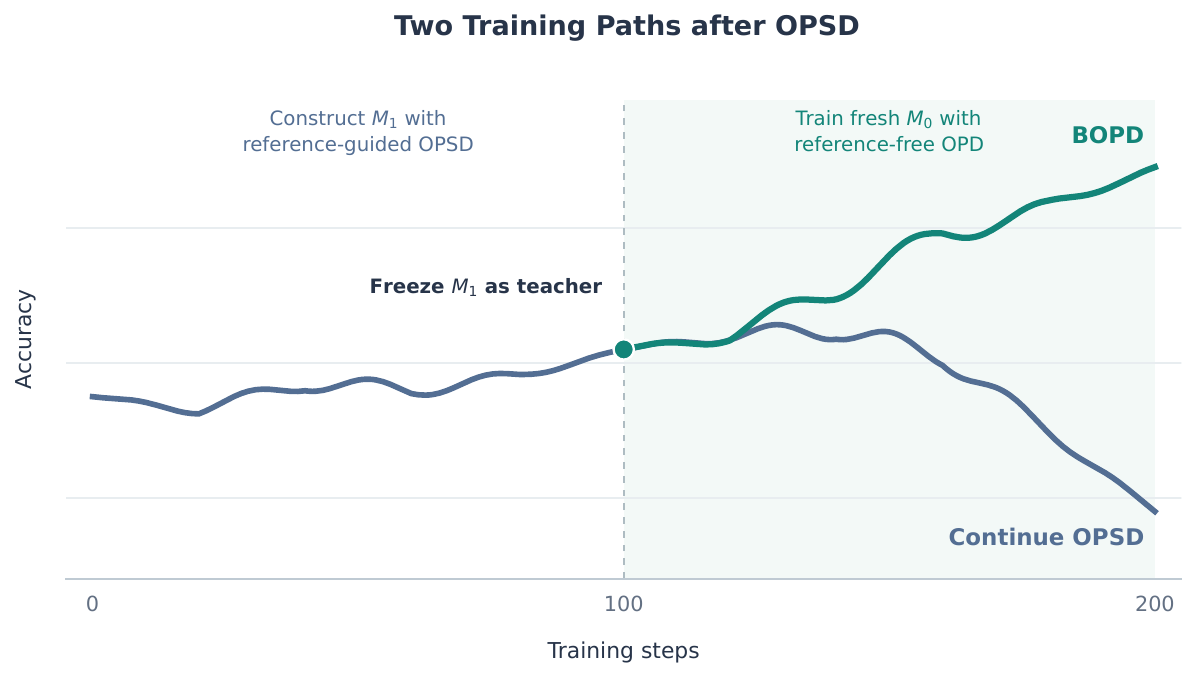}
    \caption{
        Performance of OPSD and BOPD during the training process
    }
    \label{fig:opsd_dynamics}
\end{wrapfigure}
On-policy distillation (OPD) has emerged as an effective approach to language model post-training~\citep{agarwal2024onpolicy,gu2024minillm}. 
Rather than learning from fixed demonstrations, the student model generates its own trajectories and receives dense token-level supervision from a teacher model, enabling it to learn directly from the teacher on its on-policy state distribution~\citep{agarwal2024onpolicy,zhang2026onpolicysft,yang2025qwen3,lu2025onpolicy}. 
This on-policy supervision reduces the mismatch between training-time and inference-time state distributions while providing denser token-level feedback than sparse outcome-level rewards, thereby enabling substantial capability improvements during post-training.

Recent work has sought to broaden the applicability of OPD by removing the need for an external teacher model. 
On-policy self-distillation (OPSD)~\citep{zhao2026selfdistilled} achieves this by using the model itself as the teacher: the student generates trajectories conditioned only on the problem, while a frozen copy of the same model additionally conditions on the reference solution works as a teacher model.
Building on this formulation, $\beta$-OPSD~\citep{xu2026betaopsd} derives OPSD from a KL-regularized reinforcement learning objective and introduces a tunable interpolation between the reference policy and the privileged teacher as the target distribution, while DualOPSD~\citep{chen2026dualopsd} jointly optimizes the teacher and student to adapt the supervision throughout training.

Despite these advances, we observe a striking phenomenon under prolonged OPSD training: OPSD improves model performance during the early stage of training, but these gains progressively diminish and can eventually reverse as optimization continues (Figure~\ref{fig:opsd_dynamics}).
During the early stage of training, the privileged reference provides useful corrective supervision, but as the student policy evolves, the growing distributional mismatch between the student and the frozen privileged teacher leads to misleading supervision~\citep{zhao2026more,zhang2026latent}.
This suggests a fundamental limitation of OPSD: although privileged reference information makes the teacher more informative, it can also make the resulting supervision increasingly incompatible with the reference-free student as training proceeds.In Section~3, we provide empirical analyses supporting this interpretation and following the principles in \citet{li2026rethinkingopd}, we show that the privilege-conditioned model does not constitute a well-matched teacher for the evolving student.
These observations raise a broader question: can a student construct a capable and well-matched teacher for itself without relying on an external teacher model?

\begin{figure}[t]
  \centering
  \includegraphics[width=\linewidth]{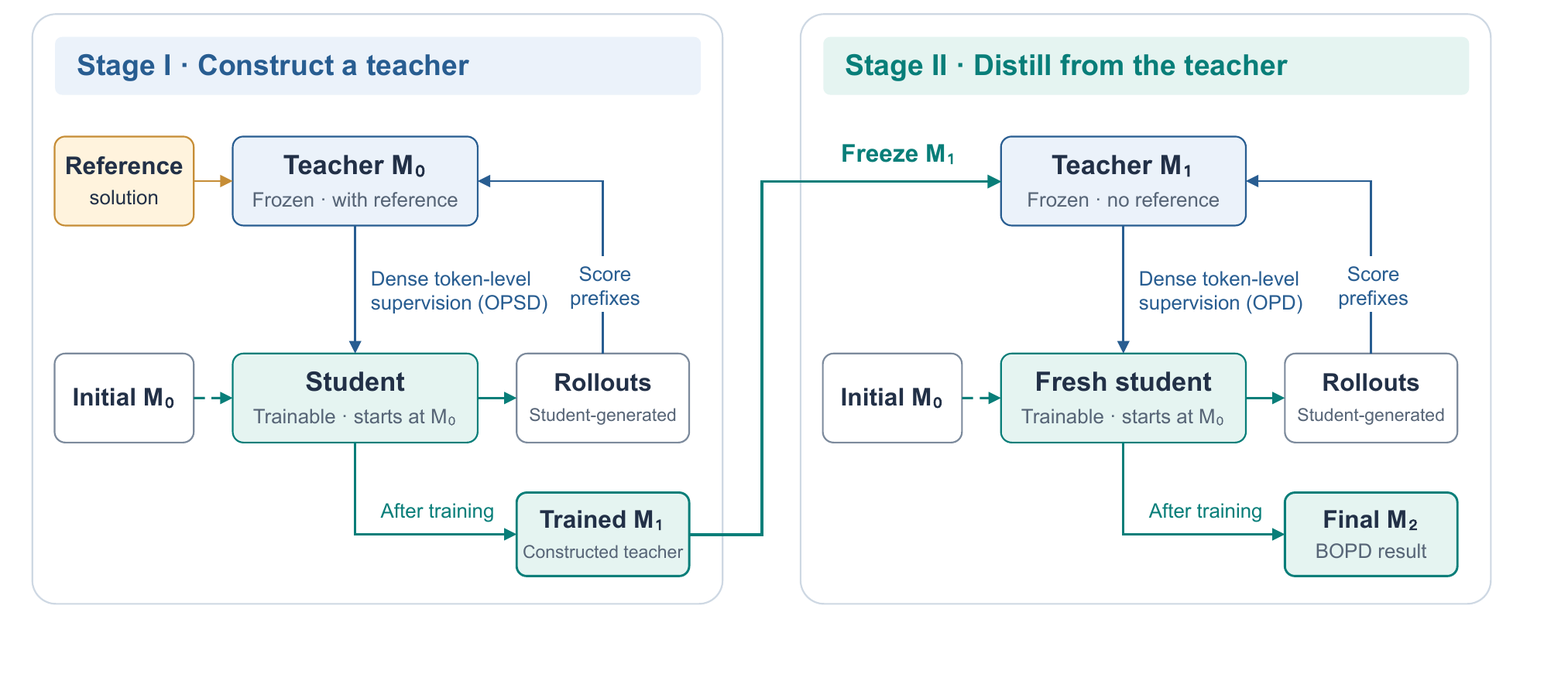}
  \caption{BOPD overview. In Stage I, a frozen copy of $M_0$ receives a reference solution and supervises a student on its own rollouts, producing $M_1$.
  In Stage II, frozen $M_1$ supervises a fresh student initialized from $M_0$ and neither model receives a reference. Dashed arrows denote student initialization; solid arrows show rollout flow, supervision, and model transfer.}
  \label{fig:bopd_overview}
  \vspace{-10pt}
\end{figure}

In this paper, we introduce Bootstrapped On-Policy Distillation (BOPD), a two-stage framework that first constructs a teacher by updating the initial model with OPSD, and then distills the resulting teacher into a fresh copy of the initial model without privileged reference conditioning.
BOPD therefore separates teacher construction from teacher use: the first stage exploits privileged reference information to internalize task-relevant reasoning signals into the model, while the second stage reuses the resulting model as a reference-free teacher for standard on-policy distillation.
Through Stage I, $\pi_{\mathrm{B}}$ absorbs part of the reasoning signal provided by the privileged reference into its own policy.
When used in Stage II, $\pi_{\mathrm{B}}$ no longer requires access to the reference solution, so the teacher and student are conditioned on the same observable information.
This design preserves the task-specific knowledge acquired through OPSD while reducing the privileged-context mismatch that arises when the teacher relies on information unavailable to the student.

Formally, the BOPD procedure can be summarized as
\begin{equation}
  \pi_0
  \xrightarrow[\text{reference available}]{\text{OPSD}}
  \pi_{\mathrm{B}}
  \quad\text{and then}\quad
  \pi_0
  \xrightarrow[\text{reference unavailable}]{\text{OPD with teacher }\pi_{\mathrm{B}}}
  \pi_{\mathrm{final}},
  \label{eq:pipeline}
\end{equation}
where \(\pi_{\mathrm{B}}\) denotes the bootstrapped teacher.
Our work makes three contributions:
\begin{itemize}
  \item We analyze the training dynamics of OPSD and provide evidence that its early improvement followed by prolonged degradation is associated with a growing incompatibility between the privileged reference-conditioned teacher and the reference-free student.

  \item We introduce Bootstrapped On-Policy Distillation (BOPD), a two-stage framework that first uses OPSD to construct a task-adapted teacher from the initial model, and then distills this teacher into a fresh copy of the initial model without privileged reference conditioning.

  \item We validate BOPD across different models through extensive experiments and ablations. BOPD improves over both the initial model and the constructed OPSD teacher, while analyses of prolonged OPSD training, teacher construction, and reference access provide empirical support for the proposed framework.
\end{itemize}

\section{Related Work}
\label{sec:related_work}

\paragraph{Learning from Reference Solutions.}
Supervised fine-tuning (SFT) learns from reference solutions by fitting their target tokens.
Recent studies have explored token weighting and selection to improve this learning signal.
DFT~\citep{wu2025dft} revisits SFT from a reinforcement-learning perspective, while PriFT~\citep{wang2026prift} and ProFit~\citep{liu2026profit} use model probabilities to determine which tokens to emphasize.
Other studies focus on modifying the training data.
\citet{zhao2025mindgap} retain correct model-generated solutions and use guided re-solving to replace incorrect ones before fine-tuning.
\citet{zhang2026onpolicysft} introduce Hinted Decoding to construct responses that better match the model's distribution.
Reference data can also be used to construct teachers.
PromptSD~\citep{ma2026promptsd} trains task-specific soft prompts on labeled examples with a frozen backbone, then uses the resulting teachers to supervise student-generated trajectories.

\paragraph{On-Policy (Self-)Distillation.}
Knowledge distillation transfers a teacher's predictive distribution to a student~\citep{hinton2015distilling}.
GKD~\citep{agarwal2024onpolicy} and MiniLLM~\citep{gu2024minillm} extend distributional supervision to student-generated trajectories, using different distillation objectives.
SKD~\citep{xu2025speculative} introduces teacher--student cooperation during trajectory generation.
\citet{li2026rethinkingopd} further study OPD and highlight the importance of teacher capability and compatibility with student-visited states.
OPSD~\citep{zhao2026selfdistilled} removes the need for a separate teacher by conditioning a self-teacher on reference solutions.
SDFT~\citep{shenfeld2026selfdistillation} uses demonstration-conditioned supervision for continual learning.
More recent studies examine the role of privileged context in distillation.
DOPD~\citep{yu2026dopd} routes token-level signals between privileged teacher and student policies, while \citet{kaur2026rethinkingopsd} identify failure modes of reference conditioning in thinking models.
LOPD~\citep{zhang2026latent} makes privileged context learnable by composing retrieved experiences into latent tokens and jointly optimizing the context composer and student.

\section{What Happens When OPSD Continues?}
\label{sec:opsd_motivation}

\begin{figure*}[t]
  \centering
 \includegraphics[width=0.95\linewidth]{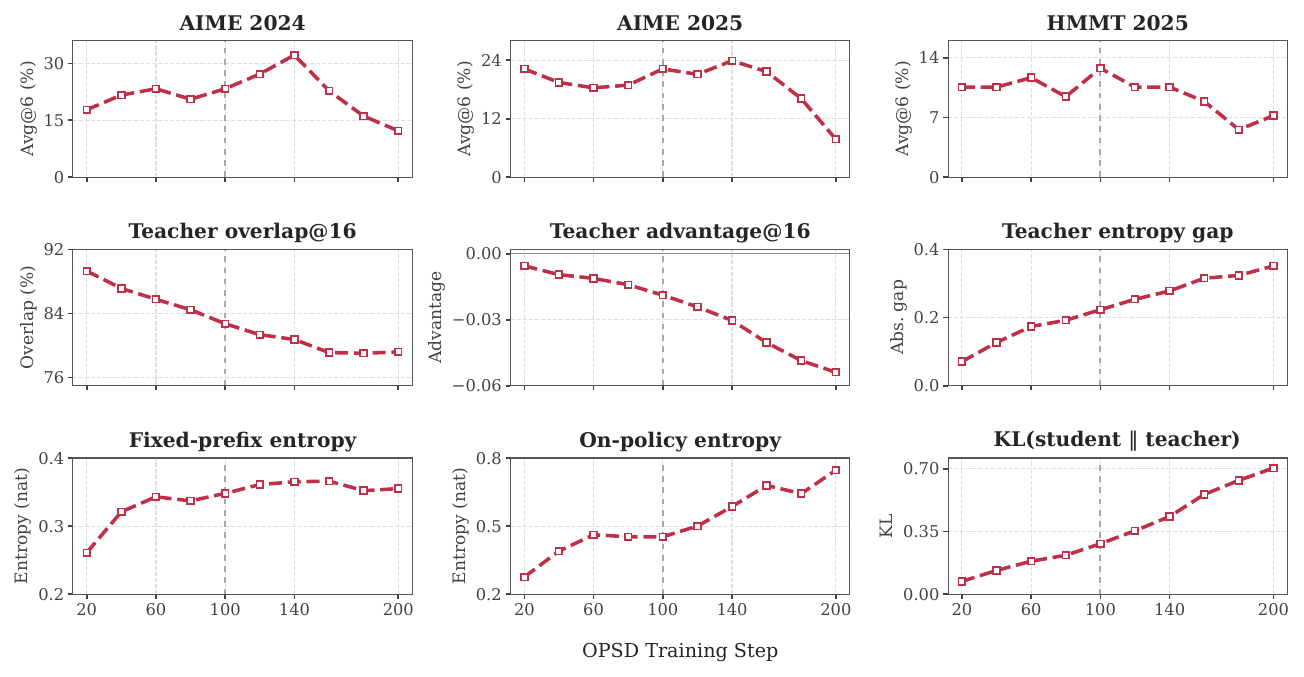}
 
  \caption{Continued OPSD training on Qwen3-4B. Top: Avg@6 on three mathematics benchmarks. Middle: overlap@16, overlap-token advantage@16, and absolute entropy gap between the OPSD student and its privileged teacher. Bottom: Fixed-prefix entropy, on-policy entropy and KL divergence between the student and teacher.
  }
  \vspace{-10pt} \label{fig:opsd_prolonged}
\end{figure*}
In this section, we study the training dynamics of OPSD beyond its initial improvement phase.
OPSD uses a frozen, privilege-conditioned initial model as the teacher to provide guidance on student-generated trajectories. 
As training proceeds, the teacher remains fixed while the student policy and the distribution of student-generated prefixes continue to evolve, causing the teacher to be queried on increasingly different states.
This raises three questions: whether the effectiveness of OPSD is sustained as training continues, whether the privileged teacher remains well matched to the evolving student, and how the student's on-policy state distribution changes over prolonged training.

To study the behavior of continued OPSD training, we train Qwen3-4B \citep{yang2025qwen3} with OPSD for 200 optimization steps and evaluate intermediate checkpoints throughout training.
We measure task performance on AIME 2024, AIME 2025, and HMMT 2025, and analyze teacher--student distributions on 64 training prompts.
For each checkpoint, we leverage the principles in \citet{li2026rethinkingopd} and track teacher--student agreement using Overlap@16, overlap-token advantage@16, the absolute entropy gap, and the full-vocabulary KL divergence.
We additionally compare student entropy on on-policy prefixes with entropy measured on prefixes generated by the initial model, allowing us to distinguish changes in the student distribution from changes in the states visited by the student during training.
Details of these measurements are provided in Appendix~\ref{sec:dynamics_details}.

\textbf{Continued OPSD does not remain consistently effective.} As shown in Figure~\ref{fig:opsd_prolonged}, task accuracy improves during the early and middle stages of training, but the gains eventually diminish and reverse. The mean Avg@6 accuracy across the three mathematical benchmarks reaches 22.22\% at step 140, but falls to 9.07\% by step 200.

\textbf{The privileged teacher is not well matched to the evolving student}.
 As shown in Figure~\ref{fig:opsd_prolonged}, teacher--student Overlap@16 decreases from 89.30\% at step 20 to 79.21\% at step 200, while the full-vocabulary KL divergence increases from 0.0719 to 0.7038. The overlap-token advantage also moves farther from zero, and the absolute entropy gap increases over training. Together, these trends show that the loss of effectiveness under continued OPSD is accompanied by growing distributional mismatch between the reference-free student and the frozen reference-conditioned teacher.

\textbf{Student uncertainty increases on evolving on-policy states.} To investigate whether this increase stems from changes in the student policy itself or from shifts in the states it visits, we compare student entropy on on-policy prefixes with entropy on a fixed set of prefixes generated by the initial model. Entropy on the fixed prefixes quickly saturates, whereas entropy on the on-policy prefixes continues to increase. This contrast suggests that the growing uncertainty is closely associated with the evolving on-policy state distribution rather than solely with changes in the student policy.

Taken together, these results suggest that continued OPSD progressively shifts the student-generated prefix distribution while the privileged teacher remains fixed, leading the teacher to provide supervision on states where its predictions are increasingly mismatched with those of the evolving student.
The observed degradation can be characterized as a growing mismatch between the privileged teacher and the reference-free student, which reduces the suitability of the teacher's supervision as training progresses. 
This motivates separating teacher construction from teacher use: rather than continuing OPSD indefinitely, we may stop at an intermediate checkpoint that has already internalized part of the task-relevant signal provided by the reference-conditioned teacher and can therefore serve as a task-adapted, reference-free teacher for a fresh student.

\section{Bootstrapped On-Policy Distillation}
\label{sec:method}

\subsection{Problem Setup and Overview}
\label{sec:problem_setup}

We consider a dataset
$\mathcal{D}=\{(x_i,y_i^\star)\}_{i=1}^{N}$, where $x_i$ is a problem and $y_i^\star$ is a reference solution. 
Let $\mathcal{D}_x$ denote the empirical prompt distribution,
$\mathcal{V}$ the token vocabulary, and
$M_0$ the initial model with parameters $\theta_0$
and policy $\pi_0=\pi_{\theta_0}$.

BOPD consists of two stages (Equation~\ref{eq:pipeline}).
In Stage I, we use OPSD to construct a task-adapted teacher
$M_1$ with policy $\pi_B=\pi_{\theta_B}$.
In Stage II, we freeze this teacher and distill it into
a fresh student initialized from $M_0$, without reference
conditioning.
Both stages operate on student-generated trajectories.
Algorithm~\ref{alg:bopd} summarizes the full procedure.

\subsection{Stage I: Bootstrapping a Teacher with OPSD}
\label{sec:stage1}

We initialize a trainable student $\pi_{\theta}$ and a frozen teacher $\pi_{\theta_0}$ from the same initial model $\pi_0$. For each $(x,y^\star)\sim\mathcal{D}$, the student generates $\hat y\sim\pi_\theta(\cdot\mid x)$. At each student-generated prefix $\hat y_{<t}$, the student and privileged teacher define
\begin{align}
  p_t(v)
  &= \pi_\theta(v\mid x,\hat y_{<t}), \\
  q_t^{\mathrm{ref}}(v)
  &= \pi_{\theta_0}(v\mid x,y^\star,\hat y_{<t}),
  \qquad v\in\mathcal{V}.
\end{align}

We update only the student parameters by minimizing the trajectory-averaged forward KL divergence:
\begin{equation}
  \mathcal{L}_{\mathrm{OPSD}}(\theta)
  =
  \mathbb{E}_{(x,y^\star)\sim\mathcal{D},\,
  \hat y\sim\pi_\theta(\cdot\mid x)}
  \left[
    \frac{1}{|\hat y|}
    \sum_{t=1}^{|\hat y|}
    D_{\mathrm{KL}}
    \!\left(
      q_t^{\mathrm{ref}}
      \middle\|
      p_t
    \right)
  \right].
  \label{eq:stage1_loss}
\end{equation}

After Stage I, we save the resulting model as the bootstrapped teacher $\pi_{\mathrm{B}}=\pi_{\theta_{\mathrm{B}}}$. 
This stage converts reference-conditioned supervision into a task-adapted model which
can subsequently be used without access to the reference solution.

\subsection{Stage II: On-Policy Distillation with the Bootstrapped Teacher}
\label{sec:stage2}

We freeze the bootstrapped teacher $\pi_B$ and
initialize a fresh student $\pi_\phi$ with
$\phi=\theta_0$.
The new student generates $\hat y\sim\pi_\phi(\cdot\mid x)$. At each prefix $\hat y_{<t}$, the student and teacher condition on the same observable context:
\begin{align}
  p_t(v)
  &= \pi_\phi(v\mid x,\hat y_{<t}), \\
  q_t^{\mathrm{B}}(v)
  &= \pi_{\mathrm{B}}(v\mid x,\hat y_{<t}).
\end{align}
Unlike Stage I, the teacher no longer receives the reference solution and we minimize
\begin{equation}
  \mathcal{L}_{\mathrm{BOPD}}(\phi)
  =
  \mathbb{E}_{x\sim\mathcal{D}_x,\,
  \hat y\sim\pi_\phi(\cdot\mid x)}
  \left[
    \frac{1}{|\hat y|}
    \sum_{t=1}^{|\hat y|}
    D_{\mathrm{KL}}
    \!\left(
      q_t^{\mathrm{B}}
      \middle\|
      p_t
    \right)
  \right].
  \label{eq:bopd}
\end{equation}

Resetting the student separates teacher construction from knowledge transfer: Stage I produces a task-adapted
teacher using privileged supervision, while Stage II transfers the resulting behavior to a fresh reference-free student on its own on-policy trajectories.

\begin{algorithm}[t]
  \caption{Bootstrapped On-Policy Distillation (BOPD)}
  \label{alg:bopd}
  \small
  \begin{algorithmic}[1]
    \Require Initial parameters $\theta_0$, reference data $\mathcal{D}$,
             stage budgets $K_1,K_2$

    \Statex \textbf{Stage I: Construct a teacher with reference-conditioned OPSD}
    \State Initialize student $\theta\gets\theta_0$ and freeze teacher $\pi_{\theta_0}$
    \For{$k=1,\ldots,K_1$}
      \State Sample a minibatch of $(x,y^\star)$ from $\mathcal{D}$
      \State Generate $\hat y\sim\pi_\theta(\cdot\mid x)$
      \State Compute $p_t(v)= \pi_\theta(v\mid x,\hat y_{<t})$ and 
      $q_t^{\mathrm{ref}}
      =\pi_{\theta_0}(\cdot\mid x,y^\star,\hat y_{<t})$
      \State Update $\theta$ by minimizing Eq.~\ref{eq:stage1_loss}
    \EndFor
    \State Set $\theta_{\mathrm B}\gets\theta$ and freeze
           $\pi_{\mathrm B}=\pi_{\theta_{\mathrm B}}$

    \Statex \textbf{Stage II: Distill from the reference-free bootstrapped teacher}
    \State Reset student $\phi\gets\theta_0$ and initialize a fresh optimizer
    \For{$k=1,\ldots,K_2$}
      \State Sample a minibatch of prompts $x$ from $\mathcal{D}_x$
      \State Generate $\hat y\sim\pi_\phi(\cdot\mid x)$
      \State Compute $p_t(v)= \pi_\phi(v\mid x,\hat y_{<t})$ and
      $q_t^{\mathrm B}
      =\pi_{\mathrm B}(\cdot\mid x,\hat y_{<t})$
      \State Update $\phi$ by minimizing Eq.~\ref{eq:bopd}
    \EndFor

    \State \Return $\pi_\phi$
  \end{algorithmic}
\end{algorithm}

\section{Experiments}
\label{sec:experiments}

\subsection{Experimental Setup}
\label{sec:setup}

\paragraph{Training data and models.}
We evaluate BOPD across different model scales and families, including Qwen3-4B, Qwen3-8B, and SmolLM3-3B~\citep{yang2025qwen3, bakouch2025smollm3}.
For mathematical reasoning experiments, we use the OpenThoughts dataset~\citep{guha2026openthoughts} and we use the TACO subset of the DeepCoder Preview Dataset for code-generation experiments~\citep{li2023taco,luo2025deepcoder}.
Within each domain, OPSD and BOPD use the same training data. We train OPSD and each stage of BOPD for 100 optimizer steps with an effective batch size of 32.
We use LoRA~\citep{hu2021lora} with rank 64 and scaling parameter 128 on the query, key, value, and output attention projections and the gate, up, and down MLP projections. For code-generation experiments, we set the dropout rate to 0.05. The learning rate is $5\times10^{-6}$, with gradient clipping at 0.1 and pointwise loss clipping at $\tau=0.05$. Student rollouts contain at most 1,024 tokens, sampled with temperature 1.1, top-p 0.95, and top-k 20. We use bfloat16 and DeepSpeed
ZeRO-2. Appendix~\ref{sec:implementation} provides additional settings.

\paragraph{Evaluation.}
For mathematical reasoning experiments, we evaluate all methods on three competition mathematics benchmarks, including AIME 2024 and AIME 2025~\citep{maaAIME}, and HMMT February 2025~\citep{hmmt2025feb}. 
We use temperature 1.0, top-p 0.8, and a 38,912-token completion limit; top-k and min-p filtering are disabled. We report average response accuracy (Avg@$K$) and the fraction of problems solved at least once (Pass@$K$), using $K = 12$ for the main results and $K = 6$ for the ablation studies. Mean accuracy is the unweighted average across the three benchmarks. 
To assess capability retention after mathematical training, we also evaluate them on HumanEval, IFEval~\citep{zhou2023ifeval}, and MMLU~\citep{hendrycks2021mmlu}.
For code-generation experiments, we evaluate all methods on HumanEval+ and MBPP+~\citep{liu2023evalplus} and the new problems in LiveCodeBench v5 and
v6~\citep{jain2025livecodebench}. We set the maximum generation length to 4,096 tokens and report Pass@1. 
Appendices~\ref{sec:metrics} and~\ref{sec:retention_details} provide further
evaluation details.

\subsection{Main Results}
\label{sec:main_results}

\begin{table}[t]
  \centering\footnotesize
  \setlength{\tabcolsep}{2pt}
  \renewcommand{\arraystretch}{1.08}
  \caption{Mathematical reasoning results (\%). 
  Bold and underlining mark the best and second-best distinct scores,
  respectively, within each model size and metric.}
  \label{tab:main_results}
  \begin{tabular*}{\linewidth}{@{\extracolsep{\fill}}cccccccccc@{}}
    \toprule
    & & \multicolumn{4}{c}{Avg@$12$ $\uparrow$} & \multicolumn{4}{c}{Pass@$12$ $\uparrow$} \\
    \cmidrule(lr){3-6}\cmidrule(l){7-10}
    Size & Method & AIME24 & AIME25 & HMMT25 & Mean & AIME24 & AIME25 & HMMT25 & Mean \\
    \midrule
    4B & Initial model & 22.50 & 19.72 & \underline{11.94} & 18.06 & 56.67 & 40.00 & \underline{30.00} & 42.22 \\
     & OPSD (Stage I) & \underline{25.83} & \underline{20.00} & \textbf{13.06} & \underline{19.63} & \underline{60.00} & \underline{46.67} & \textbf{33.33} & \underline{46.67} \\
     & \textbf{BOPD (ours)} & \textbf{34.17} & \textbf{26.39} & 11.39 & \textbf{23.98} & \textbf{66.67} & \textbf{63.33} & \textbf{33.33} & \textbf{54.44} \\
    \midrule
    8B & Initial model & 25.56 & 19.44 & \underline{11.11} & 18.70 & 43.33 & 33.33 & \underline{23.33} & 33.33 \\
     & OPSD (Stage I) & \underline{26.67} & \underline{21.67} & \textbf{12.22} & \underline{20.19} & \underline{46.67} & \underline{36.67} & \textbf{26.67} & \underline{36.67} \\
     & \textbf{BOPD (ours)} & \textbf{31.11} & \textbf{23.89} & \underline{11.11} & \textbf{22.04} & \textbf{56.67} & \textbf{40.00} & \underline{23.33} & \textbf{40.00} \\
    \bottomrule
  \end{tabular*}
\end{table}

\begin{table}[t]
  \centering\footnotesize
  \setlength{\tabcolsep}{2pt}
  \renewcommand{\arraystretch}{1.08}
  \caption{Code-generation results. Mean equally averages the four
  benchmarks before rounding. Bold and underlining mark the best and
  second-best distinct scores within each model size and metric.}
  \label{tab:code_results}
  \begin{tabular*}{\linewidth}{@{\extracolsep{\fill}}ccccccc@{}}
    \toprule
    Size & Method & HumanEval+ & MBPP+ & LCB v5-new & LCB v6-new & Mean \\
    \midrule
    4B & Initial model & \underline{74.39} & 57.14 & \underline{21.56} & \textbf{24.57} & 44.42 \\
       & OPSD (Stage I) & \textbf{75.61} & \underline{60.32} & \textbf{22.75} & \underline{23.43} & \underline{45.53} \\
       & \textbf{BOPD (ours)} & \textbf{75.61} & \textbf{63.23} & \textbf{22.75} & \textbf{24.57} & \textbf{46.54} \\
    \midrule
    8B & Initial model & \textbf{81.71} & \textbf{65.87} & \textbf{22.75} & 24.57 & \underline{48.73} \\
       & OPSD (Stage I) & \underline{79.27} & \underline{62.96} & \underline{22.16} & \underline{25.14} & 47.38 \\
       & \textbf{BOPD (ours)} & \underline{79.27} & \textbf{65.87} & \underline{22.16} & \textbf{28.00} & \textbf{48.82} \\
    \bottomrule
  \end{tabular*}
\end{table}

\paragraph{Mathematical reasoning.}
BOPD outperforms both the initial model and its constructed teacher at 4B
and 8B (Table~\ref{tab:main_results}). At 4B, average accuracy increases
from 18.06\% to 23.98\%, surpassing the OPSD teacher's 19.63\%.
At 8B, it rises from 18.70\% to 22.04\%, compared with 20.19\% for the
teacher. Pass@$12$ also improves at both sizes. These results show that a
model can construct a teacher from reference data and use it to train a
student that surpasses the teacher itself. Additional results under thinking-mode and on SmolLM3-3B further support the effectiveness of BOPD across inference settings and model families (Appendices~\ref{app:thinking_eval} and~\ref{app:smollm}). 

\paragraph{Code generation.}
Table~\ref{tab:code_results} reports results on coding benchmarks for models
trained on code.
At 4B, BOPD achieves 46.54\% mean Pass@1, compared with 44.42\% for the
initial model and 45.53\% for OPSD. At 8B, BOPD reaches 48.82\%, compared
with 48.73\% for the initial model and 47.38\% for OPSD. BOPD
outperforms its OPSD teacher in mean Pass@1 at both sizes.

\begin{figure}[t]
  \centering
  \includegraphics[width=\linewidth]{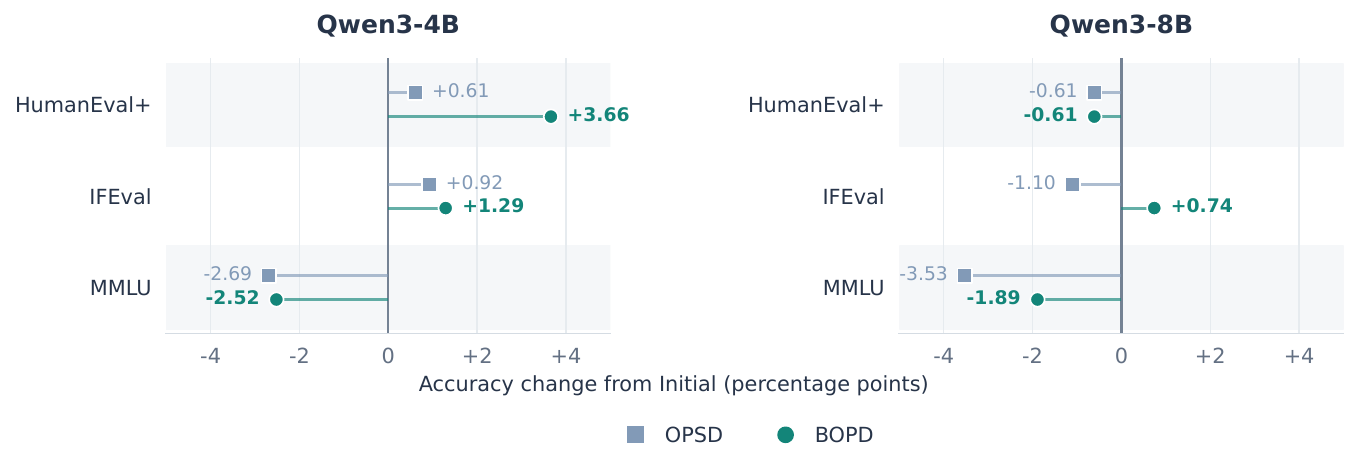}
  \caption{Capability retention after mathematical training. Points show
  accuracy changes relative to the initial model (zero line), in percentage
  points.}
  \label{fig:retention}
\end{figure}

\paragraph{Capability retention.}
Figure~\ref{fig:retention} evaluates capabilities outside mathematics after
mathematical training. On HumanEval+, BOPD scores 78.05\% versus 74.39\%
for the initial model at 4B, and 81.10\% versus 81.71\% at 8B. It raises
IFEval prompt-level strict accuracy by 1.29 and 0.74
percentage points at 4B and 8B, respectively. On MMLU, BOPD improves over
OPSD at both sizes; at 8B, it narrows
the gap according to initial model from 3.53 to 1.89 percentage points. These results show BOPD could retain more out domain ability. Appendix~\ref{sec:retention_details}
provides further evaluation details.

\begin{figure*}[t]
  \centering
  \includegraphics[width=\linewidth]{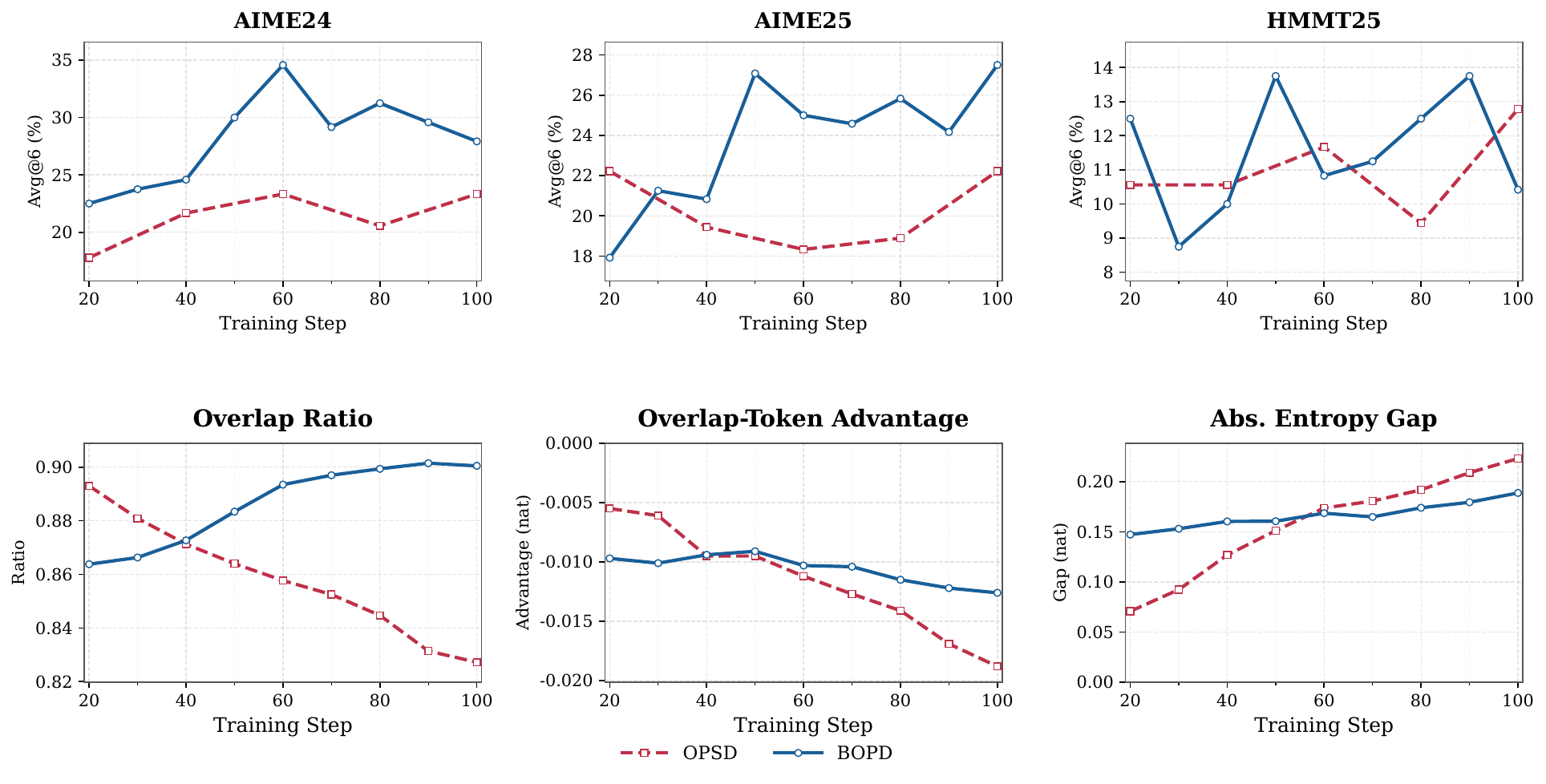}
  \caption{Accuracy and teacher--student alignment during 4B mathematical
  training. The top row shows checkpoint accuracy on each benchmark; the OPSD
  scores are also shown in Figure~\ref{fig:opsd_prolonged}. The bottom row
  reports alignment metrics on regenerated trajectories for the same 64 fixed
  training prompts. Curves begin at step 20; steps are relative to each
  method's training stage.}
  \label{fig:math_dynamics}
\end{figure*}

\subsection{Mechanism Analysis}
\label{sec:alignment_analysis}
Prior analyses of on-policy distillation suggest that an effective teacher should provide useful supervision while remaining sufficiently compatible with the student on the states visited during training. We therefore examine whether the bootstrapped teacher in BOPD exhibits a more favorable teacher–student relationship than the privileged teacher in OPSD. Following \cite{li2026rethinkingopd}, we track checkpoint accuracy together with Overlap@16, overlap-token advantage@16, and the absolute entropy gap during 4B training (Figure \ref{fig:math_dynamics}). The distributional metrics are computed on the same 64 fixed training prompts used in Section 3, with each checkpoint generating its own on-policy trajectories.

Figure \ref{fig:math_dynamics} shows a clear difference in candidate-set agreement. Under continued OPSD, Overlap@16 decreases from 0.8930 at step 20 to 0.8272 at step 100, whereas under BOPD it increases from 0.8638 to 0.9005. Thus, the BOPD student increasingly shares its teacher's likely top-16 token candidates, while the OPSD student progressively moves away from those of its privileged teacher. This trend accompanies stronger BOPD checkpoint accuracy on AIME24 and AIME25 after step 40.

The other metrics show a similar contrast. Under BOPD, the overlap-token advantage remains relatively close to zero with modest fluctuations, reaching $-0.0126$ at step 100. Under OPSD, by contrast, it moves steadily away from zero, from $-0.0055$ to $-0.0188$. A similar pattern appears in the entropy gap: BOPD shows only a moderate increase from 0.1472 to 0.1887 nats, whereas OPSD starts with a much smaller gap of 0.0707 nats but increases substantially to 0.2230 nats. Together with the opposing Overlap@16 trends, these results indicate that the BOPD teacher remains better matched to the evolving student, whereas the privileged OPSD teacher becomes increasingly mismatched as training proceeds. This growing mismatch may limit the quality of supervision provided by the OPSD teacher at later stages of training.

Taken together, these dynamics are consistent with the bootstrapped model serving as a better-matched teacher. They support the role of teacher bootstrapping in converting reference-conditioned supervision into a more compatible reference-free teacher.

\subsection{Ablation Studies}
\label{sec:ablations}

\begin{figure}[t]
  \centering
  \includegraphics[width=\linewidth]{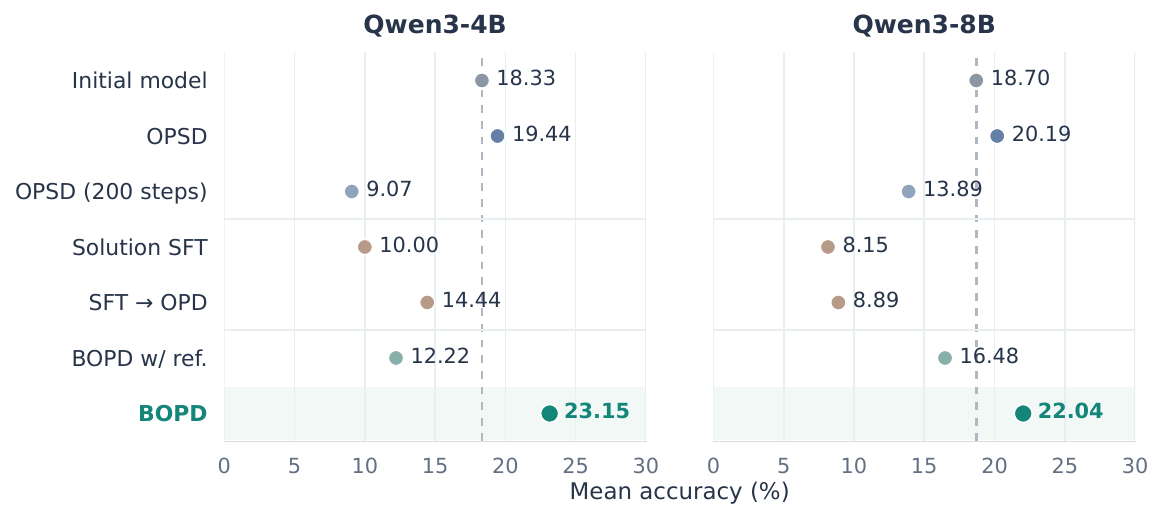}
  \caption{Mean mathematical accuracy across AIME24, AIME25, and HMMT25.
  Each gray dashed line marks the corresponding initial-model accuracy.}
  \label{fig:ablations}
\end{figure}

We ask how to construct a suitable teacher for OPD and whether BOPD makes better use
of its training steps than continued OPSD (Figure~\ref{fig:ablations}).
With the same 200-step budget, BOPD's 100 OPSD steps followed by 100 OPD
steps outperform 200 steps of OPSD at both model sizes. We also construct
a teacher with solution SFT instead of OPSD. Its accuracy is below the
initial model at both sizes. Subsequent OPD improves on this SFT teacher,
but remains below the initial model and BOPD. Finally, we give the
OPSD-trained teacher the reference solution again during Stage II. Compared
with reference-free BOPD, this
reduces average accuracy from 23.15\% to 12.22\% at 4B and from 22.04\% to
16.48\% at 8B. The trained teacher does
not need reference access to guide Stage II in this setting. One possible
explanation is that the added reference changes its targets in ways the
reference-free student cannot readily follow.
Appendix~\ref{sec:ablation_details} reports detailed scores and control settings.

\paragraph{Further distillation.}
We ask whether the BOPD student $M_2$ can support further self-improvement
(Table~\ref{tab:rounds}). First, we freeze $M_2$ as a teacher and distill it
into a fresh $M_0$ student. The resulting $M_3$ reaches 19.44\% average
accuracy, below $M_2$'s 23.15\%: a stronger student does not automatically
transfer its gains when used as a teacher. Next, we test whether $M_2$ itself
can improve through another OPSD stage. We initialize the student and frozen,
reference-conditioned teacher from $M_2$, obtaining $M_4$; its accuracy
instead falls to 3.33\%. We then complete a second BOPD cycle from $M_2$ by
using $M_4$ as a reference-free teacher for a new $M_2$ student. This yields
$M_5$ with 6.11\% accuracy. Finally, we test whether $M_4$ can teach a fresh
$M_0$ student instead, obtaining $M_6$ with 5.19\%. Neither student taught
by $M_4$ recovers $M_2$'s performance.
Appendix~\ref{sec:control_details} reports the protocol and detailed scores.

\begin{table}[t]
  \centering\small
  \setlength{\tabcolsep}{3pt}
  \caption{Further distillation at 4B. Each trained model in the first column
  is obtained by initializing from the model in the Student column and
  learning from the model in the Teacher column. ``+ ref.'' means the teacher
  also sees the reference solution.}
  \label{tab:rounds}
  \begin{tabular*}{\linewidth}{@{\extracolsep{\fill}}lllrrrrr@{}}
    \toprule
    Model & Student & Teacher & AIME24 & AIME25 & HMMT25 & Avg@6 & Pass@6 \\
    \midrule
    Initial $M_0$ & --- & --- & 24.44 & 21.11 & 9.44 & 18.33 & 33.33 \\
    OPSD $M_1$ & $M_0$ & $M_0$ + ref. & 25.00 & 19.44 & 13.89 & 19.44 & 38.89 \\
    BOPD $M_2$ & $M_0$ & $M_1$ & 32.78 & 25.56 & 11.11 & 23.15 & 44.44 \\
    \midrule
    $M_3$ & $M_0$ & $M_2$ & 29.44 & 20.00 & 8.89 & 19.44 & 37.78 \\
    $M_4$ & $M_2$ & $M_2$ + ref. & 3.89 & 4.44 & 1.67 & 3.33 & 15.56 \\
    $M_5$ & $M_2$ & $M_4$ & 8.89 & 6.11 & 3.33 & 6.11 & 21.11 \\
    $M_6$ & $M_0$ & $M_4$ & 8.89 & 5.00 & 1.67 & 5.19 & 17.78 \\
    \bottomrule
  \end{tabular*}
\end{table}

\section{Conclusion}
\label{sec:conclusion}

We study the training dynamics of on-policy self-distillation and find that its
early gains are not sustained under continued training. As the student evolves,
the privileged reference-conditioned teacher becomes increasingly mismatched
with the reference-free student on its on-policy states. Motivated by this
observation, we introduce Bootstrapped On-Policy Distillation (BOPD), which uses
OPSD to construct a task-adapted intermediate model and then reuses this model
as a reference-free teacher for a fresh student.

Across different models, BOPD improves mathematical reasoning beyond both the
initial model and the constructed OPSD teacher, and also outperforms the
constructed teacher on code generation. Ablations show that simply extending
OPSD training, constructing the teacher with solution SFT, or restoring
reference access during the second stage does not reproduce these gains under
our setting. Training-dynamics analyses further show that BOPD maintains a more
favorable teacher--student relationship than continued OPSD. At the same time,
our further-distillation experiments show that these gains do not compound
indefinitely. Overall, our results suggest that OPSD can be used effectively as
a teacher-construction mechanism, while separating privileged teacher
construction from subsequent reference-free distillation provides a practical
route to self-improvement without an external teacher model.

\section*{AI Use Statement}

Generative AI tools assisted with refining the paper narrative and
and drafting code for analysis and figures. 
The authors reviewed the AI-assisted outputs, independently checked
all reported experimental results, and take responsibility for the paper's
content, claims, and artifacts.

\section*{Acknowledgements}
The authors acknowledge support from The Hong Kong Polytechnic University (Research Grant P0046811).

\bibliography{iclr2027_conference}
\bibliographystyle{iclr2027_conference}

\clearpage
\appendix

\section{Implementation and Evaluation Details}
\label{sec:implementation}

The source dataset contains 29,434 mathematical problem--solution pairs.
The 4B and 8B controls reuse the same 3,200 examples across stages.
Each stage generates 3,200 student rollouts, for 6,400 across BOPD's two
stages.

Table~\ref{tab:training_settings} records the settings shared by the main
training runs. We use LoRA on the attention projections
\texttt{q\_proj}, \texttt{k\_proj}, \texttt{v\_proj}, and \texttt{o\_proj},
and the feed-forward projections \texttt{gate\_proj}, \texttt{up\_proj}, and
\texttt{down\_proj}. We report the step-100 checkpoint.

\begin{table}[ht]
  \centering
  \small
  \caption{Recorded training settings for the main mathematical experiments.}
  \label{tab:training_settings}
  \begin{tabular}{ll}
    \toprule
    Setting & Value \\
    \midrule
    Training examples available & 29,434 \\
    Optimizer steps per stage & 100 \\
    GPUs / batch / accumulation (4B) & 4 / 4 / 2 \\
    GPUs / batch / accumulation (8B) & 2 / 2 / 8 \\
    Effective batch size & 32 \\
    Student rollouts per stage & 3,200 \\
    LoRA rank / scaling parameter & 64 / 128 \\
    Learning rate / gradient clipping & \(5\times10^{-6}\) / 0.1 \\
    Distillation objective & Clipped forward KL \\
    Pointwise clipping threshold & 0.05 \\
    Loss parameters & \(\beta=0\), \(\lambda=1\) \\
    Rollout temperature / top-p / top-k & 1.1 / 0.95 / 20 \\
    Maximum rollout output / sequence length & 1,024 / 20,000 tokens \\
    Precision / distributed training & BF16 / DeepSpeed ZeRO-2 \\
    Attention implementation & FlashAttention 2 \\
    Rollout engine & Colocated vLLM, tensor parallelism 1 \\
    vLLM memory fraction (4B / 8B) & 0.4 / 0.35 \\
    Thinking mode (student / teacher) & Off / off \\
    Teacher updates & None; teacher frozen within each stage \\
    Stage II student initialization & Exact initial checkpoint \\
    \bottomrule
  \end{tabular}
\end{table}

\subsection{Evaluation Metrics and Protocols}
\label{sec:metrics}

For benchmark \(b\), let \(c_{i,j}\in\{0,1\}\) indicate whether response
\(j\) to question \(i\) is correct. With \(N_b=30\) and \(K\) responses
per problem, we compute
\begin{align}
  \operatorname{Avg@}K_b
  &=\frac{100}{N_bK}\sum_{i=1}^{N_b}\sum_{j=1}^{K}c_{i,j},\\
  \operatorname{Pass@}K_b
  &=\frac{100}{N_b}\sum_{i=1}^{N_b}
    \mathbb{I}\!\left[\sum_{j=1}^{K}c_{i,j}>0\right].
\end{align}
Table~\ref{tab:main_results} uses $K=12$ at 4B and 8B;
the ablation tables use $K=6$.  We average each metric over the three benchmarks for the table mean. Displayed
values are rounded to two decimal places; the mean uses the underlying scores.
The main mathematical evaluation sets \(\texttt{top\_k}=-1\), \(\texttt{min\_p}=0\), and
\(\texttt{presence\_penalty}=0\), in addition to temperature 1.0,
\(\texttt{top\_p}=0.8\), and the 38,912-token output limit. Thinking mode is
disabled for every evaluated model in the main tables.

Code-trained models use the same fixed 3,200 TACO problems for OPSD and BOPD.
HumanEval+ and MBPP+ use the official EvalPlus tests; LiveCodeBench v5-new
and v6-new contain the problems newly added relative to the preceding release.
All four code benchmarks use greedy decoding, one response per problem, a
4,096-token generation limit, an 8,192-token context limit, and seed 0.

\subsection{Thinking-Mode Evaluation}
\label{app:thinking_eval}

We evaluate OPSD and BOPD with Qwen3-4B and Qwen3-8B on AIME24, AIME25, and HMMT25, using six generations per problem. In OPSD, the student generates non-thinking rollouts, while the teacher uses thinking and is conditioned on the reference distribution. We use a clipping threshold of 0.05 for 4B and 0.06 for 8B. In BOPD, a fresh student initialized from the base model generates non-thinking rollouts and distills from the corresponding OPSD-trained thinking teacher without reference conditioning or clipping. All evaluations enable thinking and use temperature 1.0, top-$p$ 0.95, and a maximum of 38,912 generated tokens. We report Avg@6, the fraction of sampled responses that are correct, and Pass@6, the fraction of problems for which at least one of the six responses is correct.

\begin{table}[t]
  \centering\footnotesize
  \setlength{\tabcolsep}{2pt}
  \renewcommand{\arraystretch}{1.08}
  \caption{Thinking-enabled mathematical reasoning results (\%).
  Each problem is sampled 12 times. Each benchmark cell reports
  Avg@12 / Pass@12. Bold and underlining mark the best and
  second-best distinct scores, respectively, within each model size
  and metric.}
  \label{tab:thinking_opsd_bopd_n12}
  \begin{tabular*}{\linewidth}{@{\extracolsep{\fill}}cccccccccc@{}}
    \toprule
    & & \multicolumn{4}{c}{Avg@$12$ $\uparrow$}
      & \multicolumn{4}{c}{Pass@$12$ $\uparrow$} \\
    \cmidrule(lr){3-6}\cmidrule(l){7-10}
    Size & Method & AIME24 & AIME25 & HMMT25 & Mean
         & AIME24 & AIME25 & HMMT25 & Mean \\
    \midrule
    4B & Initial model
       & 73.61 & 67.50 & 44.72 & 61.94
       & 83.33 & \textbf{86.67} & 66.67 & \textbf{78.89} \\
       & OPSD (Stage I)
       & \textbf{77.22} & 64.17 & 38.61 & 60.00
       & \textbf{90.00} & 80.00 & 60.00 & 76.67 \\
       & \textbf{BOPD (ours)}
       & \underline{75.28} & \textbf{69.44} & \textbf{45.28} & \textbf{63.33}
       & 83.33 & \underline{83.33} & \textbf{66.67} & \underline{77.78} \\
    \midrule
    8B & Initial model
       & 75.83 & 68.89 & \textbf{46.67} & 63.80
       & 83.33 & 83.33 & \textbf{66.67} & 77.78 \\
       & OPSD (Stage I)
       & \underline{78.33} & 66.11 & \textbf{46.67} & 63.70
       & 86.67 & 83.33 & \textbf{66.67} & \underline{78.89} \\
       & \textbf{BOPD (ours)}
       & \textbf{80.00} & \textbf{71.39} & 45.83 & \textbf{65.74}
       & \textbf{93.33} & \textbf{86.67} & \textbf{66.67} & \textbf{82.22} \\
    \bottomrule
  \end{tabular*}
\end{table}

\section{Retention Evaluation Details}
\label{sec:retention_details}

We evaluate all 541 IFEval prompts and all 164 HumanEval problems.
Both evaluations use a 4,096-token output limit; IFEval additionally uses
a 16,384-token context limit. Between 8 and 25 IFEval responses per
configuration reach the output limit.

IFEval uses BF16, LoRA, greedy decoding, one response per prompt, and seed 0.
The official Google scorer is pinned to commit
\texttt{4700efb9afa54286b0e04473ba80a13e8461e25f}.
All responses, including truncated ones, enter the reported scores.
For code, we use EvalPlus 0.3.1 with HumanEvalPlus v0.1.10 and the full
set of enhanced tests, greedy decoding, one response per problem, and seed 0.

\begin{table}[t]
  \centering\small
  \setlength{\tabcolsep}{4pt}
  \caption{IFEval accuracy (\%) after mathematical training. P and I denote
  prompt and instruction accuracy. Trunc. counts responses reaching the output
  limit; they remain included in scores.}
  \label{tab:ifeval}
  \begin{tabular}{llrrrrr}
    \toprule
    Size & Configuration & P strict & I strict & P loose & I loose & Trunc. \\
    \midrule
    4B & Initial model & 80.41 & 86.21 & 83.73 & 88.61 & 8 \\
       & OPSD (Stage I) & 81.33 & 87.05 & 84.29 & 89.21 & 12 \\
       & Stage II OPSD & 81.70 & 87.05 & 84.84 & 89.21 & 17 \\
       & BOPD & 81.70 & 86.81 & 85.21 & 89.33 & 11 \\
       & OPSD (2$\times$ steps) & 79.30 & 85.73 & 82.81 & 88.25 & 9 \\
    \midrule
    8B & Initial model & 82.62 & 88.37 & 86.14 & 90.77 & 11 \\
       & OPSD (Stage I) & 81.52 & 87.65 & 85.77 & 90.53 & 25 \\
       & Stage II OPSD & 82.26 & 87.65 & 85.40 & 89.81 & 20 \\
       & BOPD & 83.36 & 88.49 & 85.58 & 90.05 & 13 \\
       & OPSD (2$\times$ steps) & 81.70 & 87.65 & 85.03 & 89.81 & 20 \\
    \bottomrule
  \end{tabular}
\end{table}

\begin{table}[ht]
  \centering\small
  \caption{Code pass@1 (\%) after mathematical training, with all 164
  problems included. Trunc. counts output-limit truncations.}
  \label{tab:code_retention_details}
  \begin{tabular*}{\linewidth}{@{\extracolsep{\fill}}llrrr@{}}
    \toprule
    Size & Configuration & HumanEval & HumanEval+ & Trunc. \\
    \midrule
    4B & Initial model & 83.54 & 74.39 & 2 \\
       & OPSD (Stage I) & 82.32 & 75.00 & 1 \\
       & Stage II OPSD & 84.15 & 78.66 & 1 \\
       & BOPD & 85.98 & 78.05 & 0 \\
       & OPSD (2$\times$ steps) & 82.93 & 76.22 & 2 \\
    \midrule
    8B & Initial model & 84.76 & 81.71 & 0 \\
       & OPSD (Stage I) & 84.76 & 81.10 & 1 \\
       & Stage II OPSD & 84.76 & 81.10 & 1 \\
       & BOPD & 83.54 & 81.10 & 1 \\
       & OPSD (2$\times$ steps) & 85.98 & 81.10 & 0 \\
    \bottomrule
  \end{tabular*}
\end{table}

\section{Mechanism Analysis Details}
\label{sec:dynamics_details}

For the alignment curves in Figures~\ref{fig:opsd_prolonged} and
\ref{fig:math_dynamics}, we use 64 fixed problems from the mathematical
training subset. We evaluate OPSD checkpoints at steps
10, 20, $\ldots$, 200 and BOPD checkpoints at steps 10, 20, $\ldots$, 100;
Figure~\ref{fig:math_dynamics} shows both methods through step 100.
Each checkpoint generates fresh on-policy trajectories. We score the
full-vocabulary distributions and aggregate over generated token positions.
The entropy gap averages
$|H(q_t)-H(p_t)|$ per position; it is not the absolute difference between
mean entropies. All entropies use natural logarithms.  Figure~\ref{fig:math_dynamics} plots checkpoint means without
smoothing. Each curve covers 100 steps within its own stage.

\section{Ablation Details}
\label{sec:ablation_details}

OPSD (200 steps) continues the Stage I run and evaluates that model directly.
These comparisons match nominal optimizer steps, not total FLOPs or wall time.
Solution SFT uses reference-only cross-entropy for 100 steps. The model run
uses a learning rate of $5\times10^{-6}$. SFT$\rightarrow$OPD freezes this
teacher and trains a fresh initial-model student for another 100 steps.
The SFT optimizer differs from OPSD, so this comparison evaluates the tested
recipes rather than an optimally tuned SFT teacher. The Stage II reference
ablation fixes the OPSD teacher, student initialization, loss, and step budget;
only the teacher's reference access changes.
\begin{table}[ht]
  \centering\footnotesize
  \setlength{\tabcolsep}{3pt}
  \caption{Ablation scores (\%). Each benchmark cell lists Avg@6 / Pass@6.
  Mean lists the corresponding unweighted benchmark means. Dashes mark
  benchmark-level Pass@6 scores unavailable for the separate 4B 200-step run.}
  \begin{tabular*}{\linewidth}{@{\extracolsep{\fill}}llrrrr@{}}
    \toprule
    Size & Configuration & AIME24 & AIME25 & HMMT25 & Mean \\
    \midrule
    4B & OPSD (2$\times$ steps) & 12.22 / -- & 7.78 / -- & 7.22 / -- & 9.07 / 22.22 \\
     & Solution SFT & 10.56 / 20.00 & 12.78 / 23.33 & 6.67 / 16.67 & 10.00 / 20.00 \\
     & SFT$\rightarrow$OPD & 20.56 / 43.33 & 15.00 / 36.67 & 7.78 / 23.33 & 14.44 / 34.44 \\
     & Stage II OPSD & 16.67 / 40.00 & 14.44 / 36.67 & 5.56 / 16.67 & 12.22 / 31.11 \\
    \midrule
    8B & OPSD (2$\times$ steps) & 20.00 / 36.67 & 16.67 / 43.33 & 5.00 / 16.67 & 13.89 / 32.22 \\
     & Solution SFT & 14.44 / 36.67 & 7.22 / 23.33 & 2.78 / 10.00 & 8.15 / 23.33 \\
     & SFT$\rightarrow$OPD & 12.78 / 36.67 & 11.67 / 33.33 & 2.22 / 10.00 & 8.89 / 26.67 \\
     & Stage II OPSD & 22.78 / 46.67 & 17.78 / 40.00 & 8.89 / 16.67 & 16.48 / 34.44 \\
    \bottomrule
  \end{tabular*}
\end{table}

\section{Mechanism Analysis Details}
\label{sec:dynamics_details}

\paragraph{Checkpoints and prompts.}
We select the same 64 problems from the mathematical training subset
(seed 20260922) for every diagnostic run. We evaluate OPSD checkpoints at
steps 10, 20, $\ldots$, 200 and BOPD checkpoints at steps 10, 20,
$\ldots$, 100. Figure~\ref{fig:opsd_prolonged} displays OPSD every 20 steps;
Figure~\ref{fig:math_dynamics} displays both methods through step 100.
All rollouts use non-thinking mode. The corresponding frozen teacher is
$M_0$ with access to each problem's reference solution for OPSD, and the
reference-free OPSD step-100 model for BOPD.

\paragraph{On-policy teacher--student diagnostics.}
For each checkpoint, we generate fresh responses to the 64 prompts with
that checkpoint and replay each generated prefix through both the student
and its frozen teacher. Let $p_t$ and $q_t$ denote their full-vocabulary
next-token distributions at position $t$.
Following \citet{li2026rethinkingopd}, we define the student and teacher
top-$k$ token sets as $S_t^{(p)}=\operatorname{TopK}(p_t,k)$ and
$S_t^{(q)}=\operatorname{TopK}(q_t,k)$, with $k=16$.
Let $I_t=S_t^{(p)}\cap S_t^{(q)}$. The overlap ratio is
\begin{equation}
\mathcal{M}_{\mathrm{overlap}}
=
\mathbb{E}_t\left[\frac{|I_t|}{k}\right].
\end{equation}

For overlap-token advantage, we normalize the student and teacher
probabilities over the shared token set $I_t$:
\begin{equation}
\bar p_t(v)=\frac{p_t(v)}{\sum_{u\in I_t}p_t(u)},
\qquad
\bar q_t(v)=\frac{q_t(v)}{\sum_{u\in I_t}q_t(u)},
\quad v\in I_t.
\end{equation}
The overlap-token advantage is then computed as
\begin{equation}
\mathcal{M}_{\mathrm{adv}}
=
\mathbb{E}_t\left[
\frac{1}{|I_t|}
\sum_{v\in I_t}
\bar p_t(v)
\left(\log\bar q_t(v)-\log\bar p_t(v)\right)
\right].
\end{equation}

We also compute the full-vocabulary $D_{\mathrm{KL}}(p_t\Vert q_t)$
and the absolute entropy gap $|H(p_t)-H(q_t)|$ at each position.
The diagnostic KL uses the student-to-teacher direction, whereas
the distillation objective uses the teacher-to-student direction.
Each plotted value is the mean over all evaluated token positions,
so longer responses contribute more positions. The entropy gap is
averaged after taking the absolute value at each position; it is not
the absolute difference between two mean entropies. We use natural
logarithms for entropy and KL. Although a teacher's weights are frozen,
its predictions can change across checkpoints because the
student-generated prefixes change.

\paragraph{Fixed-prefix and on-policy student entropy.}
To produce the bottom row of Figure~\ref{fig:opsd_prolonged}, we also
generate a set of reference-free $M_0$ trajectories for the same 64
prompts and hold their prefixes fixed. At OPSD step $s$, we evaluate
the student distribution $p_s(\cdot\mid u)$ at every position $u$ in
this fixed set and average its full-vocabulary entropy. For the on-policy
curve, we instead average $H(p_s(\cdot\mid u))$ over positions in that
checkpoint's own generated trajectories. Thus, the fixed-prefix curve
holds the evaluated contexts constant across checkpoints, whereas the
on-policy curve also reflects changes in the contexts visited by the
student. The two curves need not contain the same number of token positions
and are not paired position by position. The bottom-right KL curve uses
the on-policy prefixes and the reference-conditioned OPSD teacher, as above.

\paragraph{Accuracy evaluation.}
The top row of Figure~\ref{fig:opsd_prolonged} evaluates AIME 2024,
AIME 2025, and HMMT February 2025, which are separate from the 64 training
prompts. It uses six responses per problem and reports Avg@6 under the
mathematics decoding protocol of Section~\ref{sec:experiments}. We plot
checkpoint values without smoothing. Steps on the OPSD and BOPD curves
in Figure~\ref{fig:math_dynamics} are relative to their respective stages,
not a shared cumulative training axis.

\section{Evaluation on SmolLM3-3B}
\label{app:smollm}

We further evaluate BOPD on SmolLM3-3B using the same training
data and hyperparameters as the main Qwen3 experiments.
Token-level loss clipping is disabled in Stage II.
Both stages use non-thinking mode and run for 100 optimization
steps each. We evaluate on AIME24, AIME25, and HMMT25 with
12 generations per problem, following the same evaluation
protocol as the main experiments.

As shown in Table~\ref{tab:smollm}, BOPD achieves a mean
Avg@$12$ of 13.15\%, compared with 7.31\% for the initial model
and 12.31\% for OPSD. Mean Pass@$12$ also increases from
20.00\% and 33.33\% to 37.78\%, respectively.
These results suggest that the effectiveness of BOPD is
not limited to the Qwen3 model family.

\begin{table}[t]
  \centering\footnotesize
  \setlength{\tabcolsep}{2pt}
  \renewcommand{\arraystretch}{1.08}
  \caption{Mathematical reasoning results on SmolLM3-3B (\%).
  Each problem is sampled 12 times. Bold and underlining
  mark the best and second-best distinct scores, respectively,
  within each metric.}
  \label{tab:smollm}
  \begin{tabular*}{\linewidth}{@{\extracolsep{\fill}}lcccccccc@{}}
    \toprule
    & \multicolumn{4}{c}{Avg@$12$ $\uparrow$}
      & \multicolumn{4}{c}{Pass@$12$ $\uparrow$} \\
    \cmidrule(lr){2-5}\cmidrule(l){6-9}
    Method & AIME24 & AIME25 & HMMT25 & Mean
           & AIME24 & AIME25 & HMMT25 & Mean \\
    \midrule
    Initial model
      & 10.00 & 9.44 & 2.50 & 7.31
      & 30.00 & 20.00 & 10.00 & 20.00 \\
    OPSD (Stage I)
      & \underline{16.39} & \textbf{13.89}
      & \underline{6.67} & \underline{12.31}
      & \textbf{50.00} & \underline{30.00}
      & \underline{20.00} & \underline{33.33} \\
    \textbf{BOPD (ours)}
      & \textbf{17.50} & \underline{12.78}
      & \textbf{9.17} & \textbf{13.15}
      & \textbf{50.00} & \textbf{36.67}
      & \textbf{26.67} & \textbf{37.78} \\
    \bottomrule
  \end{tabular*}
\end{table}

\section{Further-Distillation Details}
\label{sec:control_details}

In Table~\ref{tab:rounds}, $M_3$ is trained from $M_0$
using $M_2$ as the teacher, while $M_4$ continues OPSD
from $M_2$ with reference-conditioned supervision.
The subsequent models $M_5$ and $M_6$ are trained
independently using the frozen $M_4$ as their teacher,
with students initialized from $M_2$ and $M_0$,
respectively.

Each distillation stage runs for 100 steps with a
fresh optimizer. All stages use the same 3,200
training examples and the same main distillation
hyperparameters. 

\end{document}